\documentclass[11pt,a4paper]{article}
\usepackage[hyperref]{emnlp-ijcnlp-2019}
\usepackage{times}
\usepackage{latexsym}
\usepackage{graphicx}
\usepackage{multirow}
\usepackage{amsmath}
\usepackage{url}
\usepackage{tikz-qtree}
\usepackage{subfig}
\usepackage{pgfplots}
\usetikzlibrary{patterns}
\usepgfplotslibrary{groupplots}

\aclfinalcopy

\title{Predicting Discourse Structure using Distant Supervision from Sentiment}

\author{Patrick Huber and Giuseppe Carenini\\
  Department of Computer Science \\
  University of British Columbia \\
  Vancouver, BC, Canada, V6T 1Z4 \\
  {\tt \{huberpat, carenini\}@cs.ubc.ca}}

\date{}

\begin{document}
\maketitle
\begin{abstract}
Discourse parsing could not yet take full advantage of the neural NLP revolution, mostly due to the lack of annotated datasets. We propose a novel approach that uses distant supervision on an auxiliary task (sentiment classification), to generate abundant data for RST-style discourse structure prediction. Our approach combines a neural variant of multiple-instance learning, using document-level supervision, with an optimal CKY-style tree generation algorithm. 
In a series of experiments, we train a discourse parser (for only structure prediction) on our automatically generated dataset and compare it with parsers trained on human-annotated corpora (news domain RST-DT and Instructional domain). Results indicate that while our parser does not yet match the performance of a parser trained and tested on the same dataset (\textit{intra-domain}), it does perform remarkably well on the much more difficult and arguably more useful task of \textit{inter-domain} discourse structure prediction, where the parser is trained on one domain and tested/applied on another one. 
\end{abstract}

\section{Introduction}
\label{Introduction}
Discourse parsing is a fundamental NLP task known to enhance key downstream tasks, such as sentiment analysis \cite{bhatia2015better,nejat2017exploring,hogenboom2015using}, text classification \cite{ji2017neural} and summarization \cite{gerani2014abstractive}.

In essence, a discourse parser should reveal the structure underlying coherent text as postulated by a discourse theory, of which the two most popular are Rhetorical Structure Theory (RST) \cite{mann1988rhetorical} and PDTB \cite{prasadpenn}. In this paper, we focus on RST-style parsing, but the proposed approach is theory agnostic and could be applied to PDTB as well. 

The RST discourse theory assumes a complete hierarchical discourse tree for a given document,
where leaf nodes are clause-like sentence fragments, called elementary-discourse-units (EDUs), while internal tree nodes are labelled with discourse relations. In addition, each node is given a nuclearity attribute, which encodes the importance of the node in its local context. 

In the past decade, traditional, probabilistic approaches, such as Support Vector Machines (SVM) \cite{hernault2010hilda, ji2014representation} and Conditional Random Fields (CRF) \cite{joty2015codra, feng2014linear}, have dominated the field. More recently, neural approaches \cite{braud2016multi, li2016discourse, braud2017cross, yu2018transition, liu2017learning, liu2018learning} have been explored, with limited success \cite{morey2017much, ferracane2019evaluating}. The main reason why recent advances in deep learning have not enhanced discourse parsing to the same extend as they have revolutionized many other areas of NLP is the small amount of training data available.
Existing corpora in English \cite{carlson2002rst, subba2009effective} only comprise of a few hundred annotated documents, each typically containing a few dozen EDUs, strictly limiting the application of deep learning methodologies. Although in principle new corpora could be created, the annotation process is expensive and time consuming. It requires sophisticated linguistic expertise and is therefore not suitable for crowd-sourcing efforts.

Another limiting issue with the available training data is the restriction to only a few domains, such as news articles \cite{carlson2002rst} or instructions \cite{subba2009effective}. This impairs the performance of existing discourse parsers when transferred into new domains.

To overcome the mentioned limitations, we propose a novel approach that uses distant supervision on the auxiliary task of sentiment classification to generate abundant data for RST-style discourse structure prediction.

\begin{figure}[t!]
\tikzset{every tree node/.style={minimum width=1.5em,draw,circle},
         blank/.style={draw=none},
         edge from parent/.style=
         {draw, edge from parent path={(\tikzparentnode) -- (\tikzchildnode)}},
         level distance=1cm, sibling distance=4.8mm}
\begin{center}
\scalebox{0.8}{
\begin{tikzpicture}
\Tree [.1-8 [.1 ] [.2-8 [.2-3 [.2 ] [.3 ] ] [.4-8 [.4 ] [.5-8 [.5-7 [.5-6 [.5 ][.6 ] ] [.7 ] ] [.8 ] ] ] ] ]
\end{tikzpicture}}
\end{center}
\caption{Example output of a strongly negative restaurant review in the Yelp'13 corpus:\newline
[Panera bread wannabes.]\textsubscript{1} [Food was okay and coffee]\textsubscript{2} [was eh.]\textsubscript{3} 
[Not large portions for the price.]\textsubscript{4} [The free chocolate chip cookie was a nice touch]\textsubscript{5}
[and the orange scone was good.]\textsubscript{6} [Broccoli cheddar soup was pretty good.]\textsubscript{7}
[I would not come back.]\textsubscript{8}
}
\label{fig:tree}
\end{figure}
We draw intuition from previous work using discourse parsing as an auxiliary task to enhance sentiment analysis \cite{ bhatia2015better,nejat2017exploring,hogenboom2015using}. Our assumption is that such synergies between sentiment analysis and discourse parsing are bidirectional. In this paper, we leverage the synergy effects in the opposite direction by using sentiment analysis to create discourse structures.

Figure 1 illustrates the discourse structure of a strongly negative Yelp review generated by our system. While solely based on sentiment information, the structure nevertheless resembles a well aligned discourse
tree. It can be observed that EDUs with negative sentiment are generally located at a higher level in the tree, while for example EDUs \textcircled{\raisebox{-0.4pt}{\small{5}}} and \textcircled{\raisebox{-0.4pt}{\small{6}}}, with positive sentiment, are at the bottom of a deep subtree. This
way, EDUs with negative sentiment strongly influence the overall sentiment, while EDUs \textcircled{\raisebox{-0.4pt}{\small{5}}} and \textcircled{\raisebox{-0.4pt}{\small{6}}} only have little impact. At the same time, semantically related EDUs generally have a shorter distance than semantically unrelated EDUs.

Our approach combines a neural variant of multiple-instance learning (MILNet) \cite{angelidis2018multiple}, with an optimal CKY-style tree generation algorithm \cite{jurafsky2014speech}. First, MILNet  computes fine-grained sentiment values and un-normalized attention scores \cite{ji2017neural} on EDU-level, by solely relying on distant supervision signals from document-level annotations. These annotations are abundantly available from several published open source datasets such as Yelp'13 \cite{tang2015document},  IMDB \cite{diao2014jointly} or Amazon \cite{zhang2015character}. Then, the sentiment values and attention scores are aggregated to guide the discourse-tree construction, optimized on the document gold-label sentiment, using optimal CKY-style parsing.

Following this approach, we generate a new corpus annotated with ``silver standard" discourse trees, which comprises of 100k documents (two orders of magnitude more than any existing corpora). To test the quality of our new corpus, we run a series of experiments, where we train the top performing discourse parser by \newcite{wang2017two} on our corpus for discourse structure prediction and  compare it  with  the same parser trained on human annotated corpora in the news domain (RST-DT) and in the instructional domain. Results indicate that while training a parser on our corpus does not yet match the performance of a parser trained and tested on the same dataset (\textit{intra-domain}), it does perform remarkably well on the significantly more difficult and arguably more useful task  of  \textit{inter-domain}  discourse  structure  prediction, where the parser is trained on one domain and tested/applied on another one. 
Our results on \textit{inter-domain} discourse parsing, shown in Section \ref{Evaluation}, strongly suggest that if anyone wants to leverage discourse parsing in a domain without annotated data, it is advantageous to use a discourse parser which has been trained on our new corpus, rather than, for instance, on RST-DT.

\section{Related Work}
\label{Related_Work}
Our approach to address the lack of annotated data in discourse parsing lies at the intersection of RST-style parsing, sentiment analysis and multiple-instance learning (MIL).

A large number of highly diverse discourse parsers have been proposed in previous work, with non-neural ones achieving the best performance. In this paper, we consider a set of top-performing parsers, which follow fundamentally different intuitions on how the parsing process should be modelled. \newcite{joty2015codra} and \newcite{ji2014representation} argue that discourse parsing should use a single model for structure, nuclearity and relation modelling. \newcite{joty2015codra} further propose to separate the task ``vertically" on sentence- and document-level, while \newcite{ji2014representation} are using a single Shift-Reduce parser based on lexical features. The current state-of-the-art system by \newcite{wang2017two} follows an opposing intuition, namely that the task should be separated ``horizontally" into two sequenced components. The first classifier models the structure and nuclearity, while the second classifier builds the relation model. 
Apart from being the state-of-the-art model, \newcite{wang2017two} has the ideal architecture for our experiments. With the horizontal separation between structure/nuclearity and relation prediction classifiers, it can be easily tailored to just make discourse structure predictions when trained on our new corpus of discourse trees.

The second related area is sentiment analysis, which we use as our auxiliary task. Previous studies, e.g., \newcite{bhatia2015better,ji2017neural}, have shown that sentiment prediction can be enhanced by leveraging discourse information, as the tree structure can influence the significance of certain clauses in the document and boost the overall performance. In particular, \newcite{bhatia2015better} use handcrafted discourse features on the sentiment classification task to score clauses depending on the level in the discourse tree. \newcite{ji2017neural} use discourse trees generated by a discourse parser \cite{ji2014representation} to inform a recursive neural network and automatically learn the model weights for sentiment prediction. In this paper, we exploit the relation between sentiment analysis and discourse parsing in the opposite direction by using sentiment annotations to create  discourse structures.

The third area of related work is distant supervision aimed at automatically generating fine-grained annotations. Distant supervision has previously been used to retrieve sentiment \cite{marchetti2012learning, tabassum2016tweetime} and emotion classes \cite{abdul2017emonet} from opinionated text, showing the potential of distant supervision with user generated content. A common technique for distant supervision is multiple-instance learning \cite{keeler1992self}, where the general idea is to retrieve fine-grained information from high-level signals. High-level signals are called \textit{bags} and fine-grained information is referred to as \textit{instances}. The task is defined as the generation of \textit{instance} labels solely based on the given \textit{bag} labels. We follow the approach by \newcite{angelidis2018multiple}, who train their MILNet system on the publicly available Yelp'13 \cite{tang2015document} and IMDB \cite{diao2014jointly} datasets. Results indicate that MILNet can capture EDU-level sentiment information as well as the relative importance of EDUs when evaluated on datasets annotated on EDU-level. In this paper, we adapt MILNet to generate information useful for deriving discourse trees from corpora with document-level sentiment annotations.

\section{Our Approach}
\label{Our_Approach}
\begin{figure*}[t!]
\begin{minipage}[t]{.48\textwidth}
  \centering
  \includegraphics[width=0.52\textwidth]{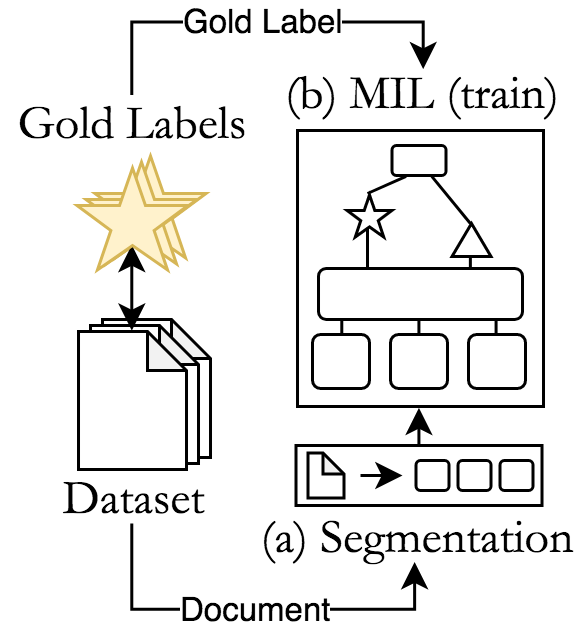}
\caption{First stage, training the MIL model on the document-level sentiment prediction task}
\label{fig:overall_1}
\end{minipage}
\hfill
\begin{minipage}[t]{.48\textwidth}
  \centering
  \includegraphics[width=1\textwidth]{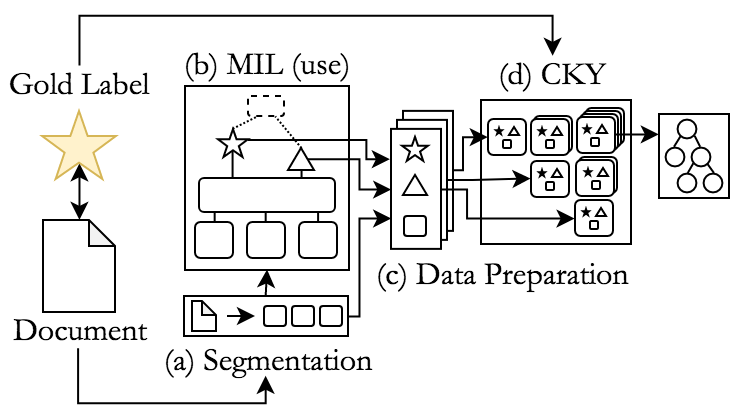}
\caption{Second stage, using the neural MIL model to retrieve fine-grained sentiment and attention scores (star/triangle), used for the CKY computation to generate the optimal discourse tree}
\label{fig:overall_2}
\end{minipage}
\end{figure*}

To generate a large number of discourse structures via distant supervision from sentiment, we propose a four-step approach, shown in Figures \ref{fig:overall_1} and \ref{fig:overall_2}. Figure \ref{fig:overall_1} illustrates the first stage of the approach, where for each document in the dataset, (a) the document is segmented into EDUs and (b) our adaptation of MILNet is trained on the document-level sentiment. Next, shown in Figure \ref{fig:overall_2}, we again (a) segment the document into EDUs and use (b) the MIL network to generate fine-grain sentiment and importance scores. Then in (c), we prepare those scores to be used in (d), the CKY-like parser, which generates an optimal RST discourse-tree for the document, based on the EDU-level scores and the gold label document sentiment.

\subsection{Segmentation and Preprocessing}
\label{seg}
We initially separate the sentiment documents into a disjoint sequence of EDUs. The segmentation is obtained using the discourse segmenter by \newcite{feng2012text} as generated and published by \newcite{angelidis2018multiple}. We preprocess the EDUs by removing infrequent- and stop-words and subsequently apply lemmatization. 

\subsection{Multiple-Instance Learning (MIL)}
\label{mil}
Our MIL model is closely related to the methodology described in \newcite{angelidis2018multiple}, as well as the papers by \newcite{yang2016hierarchical} and \newcite{ji2017neural}. The computation is based on the initial segmentation described in section \ref{seg} and is shown in further detail in Figure \ref{fig:mil}.

Our model consists of two levels of Recurrent Neural Networks (RNN) inspired by \newcite{yang2016hierarchical} and a sentiment- and attention-module. The computational flow in the model is defined from bottom to top in Figure \ref{fig:mil}. In a first step, the sparse one-hot word representations are transformed into dense vector-representations $w_i$ using a pretrained GloVe word embedding matrix \cite{pennington2014glove}. The dense word representations of a single EDU $E_i = (w_j, ..., w_k)$ are used as the sequential input for the EDU-level RNN, implemented as a bi-directional GRU module with a standard attention mechanism \cite{bahdanau2014neural}. The attention-weighted hidden-states $R_{w_i} = H_{w_i} * A_{w_i}$ are concatenated along the time axis to represent $R_{E_i}$. The second RNN on document-level subsequently uses the distributed EDU representations $(R_{E_1}, ..., R_{E_s})$ as inputs for the neural network. Based on the sequence of computed hidden-states $(H_{E_1}, ..., H_{E_s})$ in the bi-directional GRU network, two parallel model components are executed, as follows:

\begin{figure*}
\centering
\includegraphics[width=0.62\textwidth]{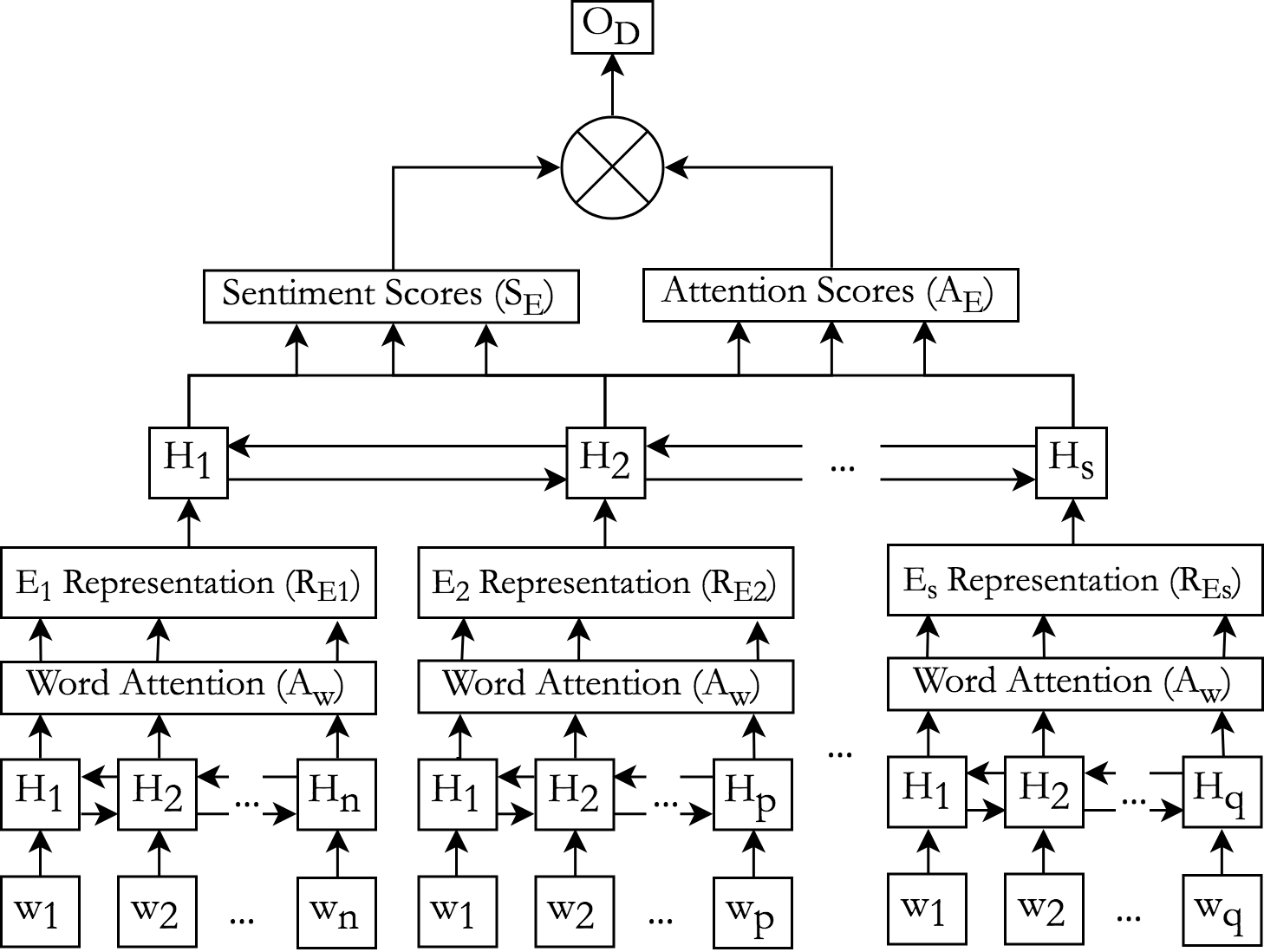}
\caption{MIL Network Topology (For readability, we leave out the second subscript of the hidden representations $H_{w_i}$ and $H_{E_i}$)}
\label{fig:mil}
\end{figure*}

\paragraph{The non-competitive attention score module} was proposed by \newcite{ji2017neural} to leverage discourse structure for sentiment prediction. By following the same intuition, we replace the softmax activation on the attention weights by a sigmoid function. This way, each attention weight $A_{E_i}$ is still limited within the range $(0,1)$, but the sum of all attention scores is not necessarily bound by $1$. 
We use the attention weight $A_{E_i}$ as the importance scores of EDU\textsubscript{i}.

\paragraph{The sentiment score module} is also executed directly on the hidden-states $(H_{E_1}, ..., H_{E_s})$ generated by the document-level RNN. To be able to interpret the dense hidden representations as sentiment predictors, we use a single feed-forward neural network layer $S$ with  $|C|$ neurons, representing the disjoint sentiment classes $(C_1, ...,C_m)$ in the dataset\footnote{In the Yelp'13 dataset, the feed-forward operation $S$ results in $5$ real-valued outputs.}. We add a sigmoid activation $sigm$ after the feed-forward layer to obtain the final, internal EDU sentiment prediction $S_{E_i} = sigm(S(H_{E_i}))$. \\

The output of the two parallel modules is multiplied EDU-wise and summed up along the time axis to calculate the final sentiment prediction of our MILNet model as $O_{D} = \sum_{E_i \in D}{S_{E_i} * A_{E_i}}$ (see top of Figure \ref{fig:mil}). 

To train our MILNet model, we use the cross-entropy loss function to compare $O_{D}$ with the gold document-level sentiment label of the review and train using the Adadelta optimizer \cite{zeiler2012adadelta}. By separating the sentiment and attention components and directly computing the final output based solely on these two values, the neural network implicitly learns the sentiment and attention scores on EDU-level as a by-product of the document-level prediction. For more information on this technique, we refer to \newcite{angelidis2018multiple}. The hyper-parameter setting of our model also mostly follows the implementation of previous work. We use a batch-size of 200 documents and train the model for 25 epochs. The bidirectional GRU layers contain 100 neurons and the model inputs are preprocessed by limiting the number of EDUs in a document to 150 and defining the maximum length of an EDU to be 20 words. With these settings, we capture over 90\% of the data by significantly decreasing the training efforts. We apply 20\% dropout on the internal sentiment layer.
\subsection{Information Extraction and Transformation}
\label{info_extraction}
Once our MILNet model is trained, we can use it to obtain the attention score $A_{E_i}$ and the sentiment score $S_{E_i}$ for each EDU $E_i$ in a document (see (c) in Figure \ref{fig:overall_2}). However, while each $A_{E_i}$ is already a scalar, $S_{E_i}$ is a vector with $|C|$ elements, one for each sentiment class $C_i$. In order to effectively combine the attention and sentiment scores for further processing, we transform $S_{E_i}$ into a scalar polarity score $pol$, centered around $0$ and uniformly distribute the $|C|$ sentiment classes within the interval of $[-1,1]$. For instance, if $|C|=5$, this would result in the five classes $pol_{\mathit{coeff}} = [-1,-0.5,0,0.5,1]$. The polarity $pol_{E_i}$ of EDU $E_i$ is computed by calculating the element-wise product of the sentiment score $S_{E_i}$ and the uniform distribution $pol_{\mathit{coeff}}$.
\begin{equation}
pol_{E_i} = \sum_{c\in C}{}S_{E_{i}}^{(c)} * pol_{\mathit{coeff}}^{(c)}
\end{equation}
We transform the gold labels in the same way to keep the representations consistent. With the polarity scores $pol_{E_i}$ replacing the original sentiment scores $S_{E_i}$, a neutral sentiment document now receives a sentiment polarity of $0$, while heavily positive or negative EDUs are mapped onto the scores $+1$ and $-1$ respectively. This way, the obtained attention scores $A_{E_i}$ and the calculated polarities $pol_{E_i}$ can be combined to create a weighted sentiment score with high attention values resulting in stronger polarities.

\subsection{CKY Tree Generation}
\label{tree_gen}
The final step in our approach (see (d) in Figure \ref{fig:overall_2}) takes the tuples of EDU-level attention scores and the generated polarities from the MILNet model to create a set of possible discourse trees. We then select the discourse tree that most precisely computes the overall sentiment of the document. To find the globally best tree, we are computing all possible tree structures (with some constraints) using a dynamic programming approach closely related to the widely used CKY algorithm \cite{jurafsky2014speech}.

To create discourse trees bottom-up using CKY, we define the necessary aggregation rules for local trees. For each binary subtree, we need to define a function $p(c_l,c_r)$ on how to aggregate the information of the two children $c_l$ and $c_r$ to represent the parent node $p$. For \textit{sentiment}, we use the intuitive attention-weighted average of the children's sentiments, defined by:
\begin{equation}
p_{s}(c_l,c_r) = \frac{ c_{l_{s}}*c_{l_{a}}+c_{r_{s}}*c_{r_{a}}}{c_{l_{a}}+c_{r_{a}}}
\end{equation}
This way, the parent sentiment does not only depend on the sentiment of its children, but also their relative importance. 

For the \textit{attention} computation we consider three different aggregation functions:

\textbf{(1)} The sum of the children's attentions (Eq. \ref{sum}). This way, the combined importance of the children is inherited by the parent node, making the node as important as the combination of all sub-nodes.
\begin{equation}
\label{sum}
p_{a_{sum}}(c_l,c_r) = (c_{l_{a}}+c_{r_{a}})*(1-\lambda)
\end{equation} 
where $\lambda$ represents a damping factor to penalize lower sub-trees, empirically chosen to be $1\%$ using grid-search. \\
\textbf{(2)} The maximum of the children's attentions. 
\begin{equation}
\label{max}
p_{a_{max}}(c_l,c_r) = max(c_{l_{a}},c_{r_{a}})
\end{equation} 
As shown in equation \ref{max}, the attention of the parent node is calculated as the maximum attention value of the two children. This aggregation function follows the intuition that the parent node is only as relevant as the most important child node. \\
\textbf{(3)} The average of the children's attentions. 
\begin{equation}
\label{avg}
p_{a_{avg}}(c_l,c_r) = \frac{c_{l_{a}},c_{r_{a}}}{2}
\end{equation} 
This aggregation function (Eq. \ref{avg}) assigns the average importance of the two children to their parent.

To create RST-style discourse documents, which can be used by existing parsers, we need to also provide nuclearity and relation labels for every tree node.
While we leave the general task of nuclearity- and relation-prediction for future work, we still need to assign those attributes to tree nodes. 
We assign nuclearity solely depending on the attention value of the children nodes, making the following binary decision: 
\begin{equation}
    c_{l_n} =
    \begin{cases}
      ``Nucleus", & \text{if}\ c_{l_{a}} \geq c_{r_{a}} \\
      ``Satellite", & \text{otherwise}
    \end{cases}
\end{equation}
This simple approach cannot assign ``Nucleus-Nucleus" nuclearity attributes, but always requires one child to be the satellite. Finally, for the necessary rhetorical relation attribute, we simply assign the \textit{span} relation to every node.

Due to the high complexity of the optimal CKY algorithm, to keep the process manageable by our computational resources\footnote{Intel Core i9-9820X, RTX 2080 Ti, 128 GB RAM}, we introduce two constraints on the generated discourse trees:
\begin{itemize}
    \item We prohibit inter-sentence relations, unless the complete sentence is represented by a single node (as shown to capture the vast majority of discourse relations by \newcite{joty2015codra})
    \item We only process documents with less or equal to 20 EDUs per document
\end{itemize}

With the aggregation functions and restrictions described above, we run the CKY-style dynamic programming approach and compare the sentiment at the root node of each of the complete discourse trees with the dataset gold-label for the document. The discourse tree with the smallest distance from the gold-label is selected as the discourse structure representation of the document (see Figure \ref{fig:overall_2}, on the right) and saved in a serializable RST-DT format. This way, we generate a dataset of 100k discourse trees.

\section{Evaluation}
\label{Evaluation}
We now describe the evaluation of our new discourse structure dataset. We start with the datasets and discourse parsers used to train and test our approach. Next, we describe the evaluation metrics, finishing with experiments and results.

\begin{table}[ht!]
\centering
\scalebox{0.85}{
\begin{tabular}{| l || r | r | r |}
\hline
Dataset & \#Documents & \#EDU/Doc & Vocab\\
\hline \hline 
Yelp'13\shortcite{tang2015document} & 335,018 & 19.1 & 183,614 \\
RST-DT\shortcite{carlson2002rst} & 385 & 56.0 & 15,503 \\
Instr-DT\shortcite{subba2009effective} & 176 & 32.6 & 3,453 \\
\hline
\end{tabular}}
\caption{Dataset size}
\label{tab:datasets}
\end{table}
\vspace{-.15in}

\subsection{Datasets}
\label{datasets}
We use three datasets to train and evaluate our approach. Table \ref{tab:datasets} summarizes the most important dataset dimensions.
\paragraph{Yelp'13} is a review dataset collected for the Yelp Dataset Challenge in 2013 \cite{tang2015document}. Every datapoint in the corpus consists of a review along with a star-rating on a 5-point scale. We use the discourse segmented version of the corpus by \newcite{angelidis2018multiple} to train our system on the auxiliary sentiment prediction task. 
\paragraph{RST-DT} is the largest and most frequently used corpus to train RST-style discourse parsers \cite{carlson2002rst}. The dataset consists of news articles from Wall Street Journal. 
We use the standard data split with 90\% training data (\textit{RST-DT\textsubscript{train}}) and 10\% test data (\textit{RST-DT\textsubscript{test}}) to test the performance of our approach against competitive baselines. 
\paragraph{Instructional Dataset}is another RST-style dataset to evaluate discourse parsers on the domain of home-repair instructions \cite{subba2009effective}. For convenience, we refer to this corpus as Instr-DT from here on. 
We separate the data into 90\% training data (\textit{Instr-DT\textsubscript{train}}) and 10\% test data (\textit{Instr-DT\textsubscript{test}}).
\paragraph{Vocabulary Overlap}is measured using the Jaccard similarity index. 
We show the absolute vocabulary sizes of the datasets in Table \ref{tab:datasets} and visualize the overlap in Table \ref{tab:overlap}.
The vocabulary overlap between the Yelp'13 corpus (containing reviews), the RST-DT dataset (on news articles) and the Instr-DT corpus (containing home-repair instructions) is predictably low, given the different domains of the datasets. While this would be a problem for models solely basing their prediction on raw input words, our system goes beyond just words as inputs. During training, we use pre-trained word embeddings to encode the inputs and the state-of-the-art discourse parser \cite{wang2017two} uses a combination of syntactical and lexical features to represent words.

\begin{table}[ht!]
\centering
\begin{tabular}{| l | r |}
\hline
Yelp'13 $\leftrightarrow$ RST-DT & 6.28\% \\
\hline
Yelp'13 $\leftrightarrow$ Instr-DT & 1.73\% \\
\hline
RST-DT $\leftrightarrow$ Instr-DT & 11.65\%\\
\hline
\end{tabular}
\caption{Vocabulary overlap between datasets}
\label{tab:overlap}
\end{table}

\subsection{Discourse Parsers}
\label{baselines}
In our experiments, we apply four simple baselines and four 
competitive discourse parsers, often used in previous work for comparison studies.
\paragraph{Right/Left Branching Baselines:} predict a binary, fully right- or left-branching tree for every document in the dataset. 
\paragraph{Hierarchical Right/Left Branching Baselines:} predict a binary, fully right- or left-branching tree on sentence-level and combine the sentence-level trees in right- or left-branching manner for every document in the dataset.
\paragraph{HILDA:} a classic, greedy, bottom-up parser using linear SVMs \cite{hernault2010hilda}.
\paragraph{DPLP:} a SVM-based shift-reduce parser build on linear projections of lexical features \cite{ji2014representation}.
\paragraph{CODRA:}  a CKY-based chart parser combined with Dynamic Conditional Random Fields, separating the computation on sentence- and document-level \cite{joty2015codra}. 
\paragraph{Two-stage Parser:} current state-of-the-art parser by \newcite{wang2017two}. Employs two separate SVM classifiers for structure/nuclearity and relations, reaching the best performance for structure and nuclearity. This is the parser we rely on in our experiments due to its performance advantage compared to other discourse parsers and its separate computation of the structure/nuclearity and the discourse relation. 
We use the publicly available code provided by \newcite{wang2017two} and remove the relation classification module.

\subsection{Metrics}
\label{metrics}
Consistent with previous work, e.g., \newcite{wang2017two, joty2015codra} and following the recent analysis by \newcite{morey2017much}, our key metric is the average micro precision on span level, computed as the global overlap of the discourse structure prediction and the gold structure. We traverse both discourse trees $tree_{pred_i}$ and $tree_{gold_i}$ of each document $i$ in post-order and compute: 
\begin{equation}
    precision = \frac{\sum_{i}{tree_{pred_i} \cap tree_{gold_i}}}{\sum_{i}{|tree_{gold_i}|}}
\end{equation}
Notice that the choice of precision over recall and F-score has no impact on the results when using manual segmentation, as shown in previous work, e.g., \newcite{wang2017two, joty2015codra}. 

\subsection{Experiments and Results}
\label{results}
\begin{figure*}[t!]
\begin{minipage}[t]{.48\textwidth}
  \centering
  \begin{tikzpicture}
  \begin{axis}[name=small,
    width=0.28*\textwidth, 
    height=5.5cm, 
    scale only axis,
    anchor=above north east,
    xlabel={Dataset Size},
    ymax = 87,
    minor tick num=1,
    ymin=55,
    yticklabel=\pgfmathparse{\tick}\pgfmathprintnumber{\pgfmathresult}\,\%,
    ylabel={Precision},
    ytick distance=5,
    xtick=data,
    xticklabels={1,2,4,8},
    ymajorgrids=true,
    xmajorgrids=true,
    ]
    \addplot
        coordinates {
        	(0,57.60)
        	(1,66.56)
        	(2,73.08)
        	(3,74.58)
        	};
        	
    \addplot
        coordinates {
        	(0,58.97)
        	(1,63.44)
        	(2,72.19)
        	(3,74.45)
        	};
        	
    \addplot
        coordinates {
        	(0,59.22)
        	(1,68.57)
        	(2,70.83)
        	(3,73.76)
        	};
        	
    \addplot[mark=triangle*,color=black]
        coordinates {
        	(0,72.31)
        	(1,76.87)
        	(2,79.46)
        	(3,80.71)
        	};
        	
    \addplot[mark=diamond*,color=purple]
        coordinates {
        	(0,57.21)
        	(1,65.04)
        	(2,66.90)
        	(3,70.72)
        	};
    \end{axis}
  \begin{axis}[name=large,
    width=0.48*\textwidth, 
    height=5.5cm, 
    scale only axis,
    at={(0.015\linewidth,0)},
    anchor=above north west,
    yticklabel pos=right,
    xlabel={Dataset Size},
    ytick distance=5,
    yticklabels={},
    ymax = 87,
    minor tick num=1,
    ymin=55,
    xtick=data,
    xticklabels={10,100*, 1k, 10k, 100k},
    legend pos=south east,
    legend style={font=\small},
    ymajorgrids=true,
    xmajorgrids=true,
]
    \addplot
        coordinates {
        	(0,76.23)
        	(1,76.30)
        	(2,76.23)
        	(3,76.25)
        	(4,76.35)
        	};
    \addlegendentry{Our\textsubscript{avg}$\rightarrow$RST-DT\textsubscript{train}}
    
    \addplot
        coordinates {
        	(0,75.35)
        	(1,76.31)
        	(2,75.84)
        	(3,76.57)
        	(4,77.12)
        	};
    \addlegendentry{Our\textsubscript{max}$\rightarrow$RST-DT\textsubscript{train}}
    
    \addplot
        coordinates {
        	(0,74.14)
        	(1,74.53)
        	(2,74.65)
        	(3,75.45)
        	(4,75.79)
        	};
    \addlegendentry{Our\textsubscript{sum}$\rightarrow$RST-DT\textsubscript{train}}
    
    \addplot[mark=triangle*,color=black]
        coordinates {
        	(0,81.24)
        	(1,85.37)
        	};
    \addlegendentry{RST-DT\textsubscript{train}$\rightarrow$RST-DT\textsubscript{test}}
    
    \addplot[mark=diamond*,color=purple]
        coordinates {
        	(0,71.08)
        	(1,74.30)
        	};
    \addlegendentry{Instr-DT\textsubscript{train}$\rightarrow$RST-DT\textsubscript{train}}
    \end{axis}    
\end{tikzpicture}
\caption{Results of training and testing on the datasets listed in the legend (*Complete dataset was used for RST-DT(385 documents) and Instr-DT(176 documents))}
\label{fig:rst_results}
\end{minipage}
\hfill
\begin{minipage}[t]{.48\textwidth}
  \centering
  \begin{tikzpicture}
  \begin{axis}[name=small,
    width=0.28*\textwidth, 
    height=5.5cm, 
    scale only axis,
    anchor=above north east,
    xlabel={Dataset Size},
    ymax = 87,
    minor tick num=1,
    ymin=55,
    yticklabel=\pgfmathparse{\tick}\pgfmathprintnumber{\pgfmathresult}\,\%,    
    ylabel={Precision},
    xtick=data,
    ytick distance=5,
    xticklabels={1,2,4,8},
    ymajorgrids=true,
    xmajorgrids=true,
    ]
    \addplot
        coordinates {
        	(0,66.01)
        	(1,67.66)
        	(2,71.48)
        	(3,72.82)
        	};
        	
    \addplot
        coordinates {
        	(0,66.28)
        	(1,67.52)
        	(2,69.24)
        	(3,71.63)
        	};
        	
    \addplot
        coordinates {
        	(0,66.63)
        	(1,67.88)
        	(2,71.12)
        	(3,71.36)
        	};
        
    \addplot[mark=diamond*,color=black]
        coordinates {
        	(0,65.27)
        	(1,69.04)
        	(2,73.74)
        	(3,74.14)
        	};
        	
    \addplot[mark=triangle*,color=purple]
        coordinates {
        	(0,69.83)
        	(1,71.50)
        	(2,72.55)
        	(3,73.25)
        	};
    \end{axis}
  \begin{axis}[name=large,
    width=0.48*\textwidth, 
    height=5.5cm, 
    scale only axis,
    at={(0.015\linewidth,0)},
    anchor=above north west,
    yticklabel pos=right,
    xlabel={Dataset Size},
    ytick distance=5,
    yticklabels={},
    ymax = 87,
    minor tick num=1,
    ymin=55,
    xtick=data,
    xticklabels={10,100*, 1k, 10k, 100k},
    legend pos=south east,
    legend style={font=\small},
    ymajorgrids=true,
    xmajorgrids=true,
]
    \addplot
        coordinates {
        	(0,72.04)
        	(1,72.11)
        	(2,73.21)
        	(3,73.87)
        	(4,74.84)
        	};
    \addlegendentry{Our\textsubscript{avg}$\rightarrow$Instr-DT\textsubscript{train}}
    
    \addplot
        coordinates {
        	(0,71.37)
        	(1,71.82)
        	(2,72.38)
        	(3,72.67)
        	(4,73.14)
        	};
    \addlegendentry{Our\textsubscript{max}$\rightarrow$Instr-DT\textsubscript{train}}
    
    \addplot
        coordinates {
        	(0,71.10)
        	(1,71.45)
        	(2,71.46)
        	(3,72.04)
        	(4,72.72)
        	};
    \addlegendentry{Our\textsubscript{sum}$\rightarrow$Instr-DT\textsubscript{train}}
    
    \addplot[mark=diamond*,color=black]
        coordinates {
        	(0,75.19)
        	(1,77.28)
        	};
    \addlegendentry{Instr-DT\textsubscript{train}$\rightarrow$Instr-DT\textsubscript{test}}
    
    \addplot[mark=triangle*,color=purple]
        coordinates {
        	(0,73.92)
        	(1,74.37)
        	};
    \addlegendentry{RST-DT\textsubscript{train}$\rightarrow$Instr-DT\textsubscript{train}}
    \end{axis}    
\end{tikzpicture}
\caption{Results of training and testing on the datasets listed in the legend (*Complete dataset was used for RST-DT(385 documents) and Instr-DT(176 documents))}
\label{fig:inst_results}
\end{minipage}
\end{figure*}

We run experiments in two phases. In the first phase, the state-of-the-art discourse parser by \newcite{wang2017two} is individually trained on each of the datasets and tested on the two corpora containing gold discourse annotations. In the second phase, the best results are placed in the broader context of competitive discourse parsers.
\paragraph{Phase 1:} We train the state-of-the-art discourse parser on five different corpora and perform tests on two corpora. The five training corpora are:  RST-DT\textsubscript{train}, Instr-DT\textsubscript{train} and the three versions of our novel dataset, generated using the different attention aggregation functions $(avg, max, sum)$ discussed in Section \ref{tree_gen}. The two corpora used for testing are RST-DT\textsubscript{train} and Instr-DT\textsubscript{train} (for which gold standard annotations are available). Notice that whenever training and testing are performed on the same corpus, the model is trained on the training portion of the dataset (RST-DT\textsubscript{train} or Instr-DT\textsubscript{train}) and evaluated on the test data (RST-DT\textsubscript{test} or Instr-DT\textsubscript{test}). Finally, since one of the key benefits of our approach is the ability to generate large dataset, we also assess the relation between the dataset size and the parser performance in this phase.
The results of all our experiments in the first phase are shown in Figure \ref{fig:rst_results} and Figure \ref{fig:inst_results}.

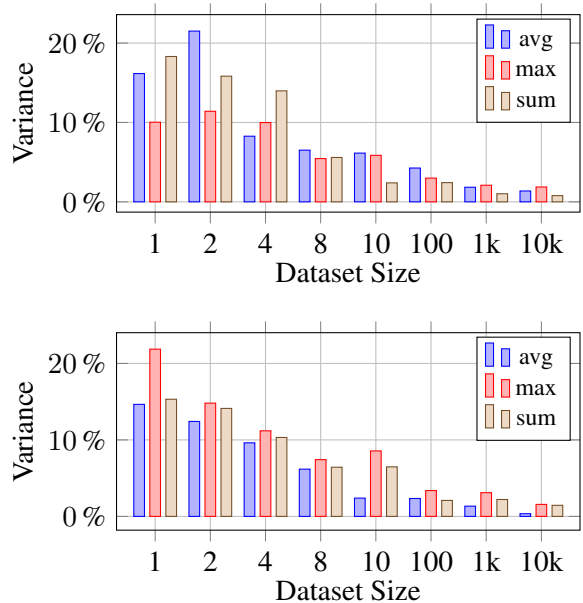
\begin{figure}[ht!]
\subfloat{
\begin{tikzpicture}
\begin{axis}[
    ybar,
	bar width=4pt,
	height=4.2cm,
    width=0.48*\textwidth, 
    ylabel={Variance},
    xlabel={Dataset Size},
    legend style={font=\small},
    yticklabel=\pgfmathparse{\tick}\pgfmathprintnumber{\pgfmathresult}\,\%,    
    xtick=data,
    xticklabels={1, 2, 4, 8, 10, 100, 1k, 10k},
    ymajorgrids=true,
    xmajorgrids=true,
    ]
    \addplot+[ybar] plot 
    coordinates {
    (0,16.16)
    (1,21.51)
    (2,8.27)
    (3,6.51)
    (4,6.14)
    (5,4.26)
    (6,1.84)
    (7,1.37)
    };
    \addlegendentry{avg}

    \addplot+[ybar] plot 
    coordinates {
    (0,10.03)
    (1,11.41)
    (2,9.99)
    (3,5.45)
    (4,5.86)
    (5,2.99)
    (6,2.09)
    (7,1.87)
    };
    \addlegendentry{max}
    
    \addplot+[ybar] plot 
    coordinates {
    (0,18.31)
    (1,15.83)
    (2,13.98)
    (3,5.59)
    (4,2.39)
    (5,2.42)
    (6,1.02)
    (7,0.78)
    };
    \addlegendentry{sum}
\end{axis}
\end{tikzpicture}}
\newline
\subfloat{
\begin{tikzpicture}
\begin{axis}[
    ybar,
	bar width=4pt,
	height=4.2cm,
    width=0.48*\textwidth, 
    ylabel={Variance},
    xlabel={Dataset Size},
    legend style={font=\small},
    yticklabel=\pgfmathparse{\tick}\pgfmathprintnumber{\pgfmathresult}\,\%,    
    xtick=data,
    xticklabels={1, 2, 4, 8, 10, 100, 1k, 10k},
    ymajorgrids=true,
    xmajorgrids=true,
    ]
    \addplot+[ybar] plot 
    coordinates {
    (0,14.65)
    (1,12.42)
    (2,9.62)
    (3,6.18)
    (4,2.39)
    (5,2.34)
    (6,1.34)
    (7,0.35)
    };
    \addlegendentry{avg}

    \addplot+[ybar] plot 
    coordinates {
    (0,21.87)
    (1,14.81)
    (2,11.19)
    (3,7.42)
    (4,8.56)
    (5,3.38)
    (6,3.10)
    (7,1.57)
    };
    \addlegendentry{max}
    
    \addplot+[ybar] plot 
    coordinates {
    (0,15.32)
    (1,14.12)
    (2,10.32)
    (3,6.43)
    (4,6.48)
    (5,2.09)
    (6,2.21)
    (7,1.45)
    };
    \addlegendentry{sum}
\end{axis}
\end{tikzpicture}
}
\caption{Sample variance across different subset sizes on the RST-DT dataset (top) and Instr-DT corpus (bottom)}
\label{fig:variance}
\end{figure}

In both figures, the left side shows the performance using small subsets of sizes $2^n$ for $n=0,1,..$ of the training data, while the right side shows the performance on large subsets of sizes $10^n$, as well as the full datasets. The precision value displayed for each subset is the average of 10 randomly selected samples from the full corpus. Figure \ref{fig:variance} shows the variance within the 10 samples, highlighting the increasing reliability of larger subsets\footnote{The complete dataset has not been subsampled and therefore does not have a variance defined.}.

The results shown in the two figures reveal several important findings:\\
\textbf{(1)}  While training the parser on our corpus does not yet match the performance of the parser trained and tested on the same dataset (\textit{intra-domain}, see black lines in Figure \ref{fig:rst_results} and \ref{fig:inst_results}), it does achieve the best performance in the \textit{inter-domain} discourse structure prediction.\\
\textbf{(2)} The performance generally increases with more training data. Larger datasets could therefore further increase the performance.\\
\textbf{(3)} Tested on the full \textit{inter-domain} training datasets (100k), the $avg$ attention-aggregation function achieves the most consistent performance on the corpora. When evaluated only on RST-DT, the $max$ aggregation function reaches the overall best performance, while the $avg$ attention-aggregation function reaches the best performance when only evaluating on Instr-DT.\\
\textbf{(4)} Small subsets of training data (4, 8 documents) already achieve relatively good results, especially on Instr-DT. One possible explanation for this behaviour is the large number of training datapoints within a single document, which generates $2*n-1$ binary training instances, where $n$ is the number of EDUs in the document.\\
\textbf{(5)} Generally, performance is highly dependent on the dataset domain and structure. In particular, the performance on the Instr-DT is generally lower and  saturates earlier than on RST-DT. Both effects could result from Instr-DT containing less and shorter documents than RST-DT.
\paragraph{Phase 2:} We further analyze our findings with respect to baselines and existing discourse parsers. The first set of results in Table \ref{tab:final} shows that the hierarchical right/left branching baselines dominate the completely right/left branching ones. However, their performance is still significantly worse than any discourse parser (\textit{intra-} and \textit{inter-domain}). 

\begin{table}[ht!]
\centering
\scalebox{0.9}{
\begin{tabular}{|l|r r|}
\hline
Approach & RST-DT\textsubscript{test} & Instr-DT\textsubscript{test}\\
\hline \hline 
Right Branching & 54.64 & 58.47 \\
Left Branching & 53.73 & 48.15 \\
Hier. Right Branch. & \textbf{70.82} & \textbf{67.86} \\
Hier. Left Branch. & 70.58 & 63.49 \\
\hline \hline
\multicolumn{3}{|c|}{\textbf{Intra-Domain} Evaluation}\\
\hline
HILDA\shortcite{hernault2010hilda} & 83.00 & --- \\
DPLP\shortcite{ji2014representation} & 82.08 & --- \\
CODRA\shortcite{joty2015codra} & 83.84 & \textbf{82.88}  \\
Two-Stage\shortcite{wang2017two} & \textbf{86.00} & 77.28 \\
\hline \hline
\multicolumn{3}{|c|}{\textbf{Inter-Domain} Evaluation}\\
\hline
Two-Stage\textsubscript{RST-DT} & $\times$ & 73.65 \\
Two-Stage\textsubscript{Instr-DT} & 74.48 & $\times$ \\
Two-Stage\textsubscript{Ours(avg)} & \underline{76.42} & \underline{\textbf{74.22}} \\
Two-Stage\textsubscript{Ours(max)} & \textbf{77.24} & 73.12 \\
\hline \hline 
Human \shortcite{morey2017much} & 88.30 & --- \\
\hline
\end{tabular}}
\caption{Discourse structure prediction results; tested on RST-DT\textsubscript{test} and Instr-DT\textsubscript{test}. Subscripts in \textit{inter-domain} evaluation sub-table indicate the training set. Best performance in the category is \textbf{bold}. Consistently best model for \textit{inter-domain} discourse structure prediction is \underline{underlined}}
\label{tab:final}
\end{table}

The second set of results show the performance of existing discourse parsers when trained and tested on the same dataset (\textit{intra-domain}). We use the results published in the original paper whenever possible. The Two-Stage approach by \newcite{wang2017two} achieves the best performance with 86\% on the structure prediction using the RST-DT dataset. On the Instructional dataset, the CODRA discourse parser by \newcite{joty2015codra} achieves the highest score with 82.88\%. 

The third set in the table shows the key results from Phase 1 on the \textit{inter-domain} performance.

Our models, learning the discourse structure solely from the \textit{inter-domain} Yelp'13 review dataset through distant supervision, reach better performance than the human annotated datasets (RST-DT and Instr-DT) when trained \textit{inter-domain}, despite the low vocabulary overlap between the Yelp'13 corpus and RST-DT/Instr-DT (Table \ref{tab:overlap}). 
While the $avg$ attention-aggregation function achieves the most consistent performance on both evaluation corpora, the $max$ function should not be dismissed, as it performs better on the larger RST-DT dataset, which is arguably more related to the Yelp'13 corpus than the sentiment-neutral Instr-DT. Furthermore, the fact that our best models reach a performance of only 8.76\% and 8.66\% below the best \textit{intra-domain} performances (tested on RST-DT and Instr-DT, respectively), shows the potential of our approach, even when compared to \textit{intra-domain} results.

\section{Conclusions and Future Work}
\label{Conclusions and Future Work}
In this paper, we address a key limitation to further progress in discourse parsing: the lack of annotated datasets. We show promising initial results to overcome this limitation by creating a large-scale dataset using distant supervision on the auxiliary task of sentiment analysis. Experiments indicate that a parser trained on our new dataset outperforms parsers trained on human annotated datasets on the challenging and very useful task of \textit{inter-domain} discourse structure prediction.

There are several directions for future work. First, given that we can now create large datasets, we intend to experiment on structure prediction with neural discourse parsers, which so far have delivered rather disappointing results. Second, an obvious next step is working on the integration of nuclearity and relation prediction to create complete RST annotations for documents from auxiliary tasks and to extend our evaluations \cite{Zeldes2017}. Third, we will study synergies between discourse parsing and further auxiliary tasks, eventually creating a single, joint system to generate globally high-quality discourse trees. Finally, instead of creating discourse structures and training existing discourse parsers on the data, we will design and implement an end-to-end system to train the complete process holistically.

\section*{Acknowledgments}
We want to thank the anonymous reviewers for their insightful comments and suggestions. This research was supported by the
Language \& Speech Innovation Lab of Cloud BU, Huawei Technologies Co., Ltd.

\bibliography{emnlp-ijcnlp-2019}
\bibliographystyle{acl_natbib}

\end{document}